\documentclass[10pt,twocolumn,letterpaper]{article}

\usepackage[final]{cvpr}      
\usepackage[accsupp]{axessibility}  
\usepackage{stfloats}
\usepackage{multirow}
\usepackage{tabularx}
\usepackage{booktabs}
\usepackage{algorithm}
\usepackage{algpseudocode}
\usepackage{booktabs} 
\usepackage{makecell} 
\usepackage{fontawesome5}
\definecolor{cvprblue}{rgb}{0.21,0.49,0.74}
\usepackage[pagebackref,breaklinks,colorlinks,allcolors=cvprblue]{hyperref}

\def\paperID{} 
\def\confName{CVPR}
\def\confYear{2026}

\title{RC-NF: Robot-Conditioned Normalizing Flow for Real-Time Anomaly Detection in Robotic Manipulation}

\author{
Shijie Zhou$^{1,2}$ \quad Bin Zhu$^{3}$ \quad Jiarui Yang$^{1,2}$ \quad Xiangyu Zhao$^{1,2}$ \quad Jingjing Chen$^{1,2}$\footnotemark \quad Yu-Gang Jiang$^{1,2}$ \\
$^{1}$Institute of Trustworthy Embodied AI, Fudan University \\
$^{2}$Shanghai Key Laboratory of Multimodal Embodied AI \quad $^{3}$Singapore Management University \\
{\tt\small {\small \href{https://heikaishuizz.github.io/RC-NF/}{\faGithub\ https://heikaishuizz.github.io/RC-NF/}} \{zhousj24, jryang24, zhaoxy25\}@m.fudan.edu.cn,} \\
{\tt\small binzhu@smu.edu.sg, \{chenjingjing, ygj\}@fudan.edu.cn} \\
\vspace{-6ex}
}
\begin{document}
\twocolumn[{%
\renewcommand\twocolumn[1][]{#1}%
\maketitle
\includegraphics[width=\textwidth]{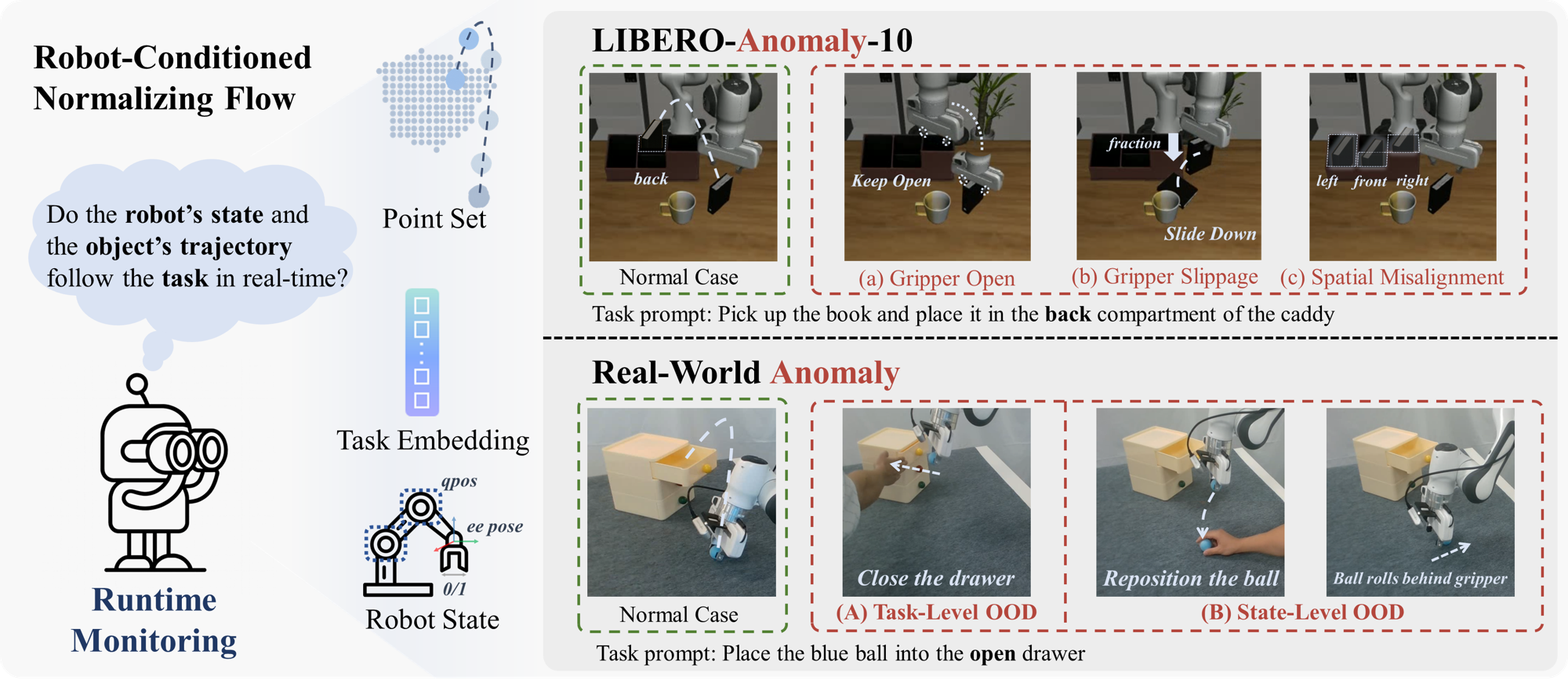}
\vspace{-4ex}
\captionof{figure}{We propose Robot-Conditioned Normalizing Flow (RC-NF) model to monitor in real time whether the robot's execution state and object motion trajectory remain consistent with the task. To evaluate anomaly (i.e., Out-of-Distribution, OOD) detection performance, we introduce LIBERO-Anomaly-10, which includes three common robotic anomalies. Real-world experiments further demonstrate RC-NF manages to enhance the adaptability of the VLA models (e.g., \( \pi_0 \)).}

\label{fig:main}
}]
\footnotetext{* Corresponding author.}
\begin{abstract}
Recent advances in Vision-Language-Action (VLA) models have enabled robots to execute increasingly complex tasks. However, VLA models trained through imitation learning struggle to operate reliably in dynamic environments and often fail under Out-of-Distribution (OOD) conditions.
To address this issue, we propose \textbf{R}obot-\textbf{C}onditioned \textbf{N}ormalizing \textbf{F}low (\textbf{RC-NF}), a real-time monitoring model for robotic anomaly detection and intervention that ensures the robot's state and the object's motion trajectory align with the task. RC-NF decouples the processing of task-aware robot and object states within the normalizing flow. It requires only positive samples for unsupervised training and calculates accurate robotic anomaly scores during inference through the probability density function.
We further present LIBERO-Anomaly-10, a benchmark comprising three categories of robotic anomalies for simulation evaluation. RC-NF achieves state-of-the-art performance across all anomaly types compared to previous methods in monitoring robotic tasks. Real-world experiments demonstrate that RC-NF operates as a plug-and-play module for VLA models (e.g., \( \pi_0 \)), providing a real-time OOD signal that enables state-level rollback or task-level replanning when necessary, with a response latency under 100 ms. These results have demonstrated that our RC-NF noticeably enhances the robustness and adaptability of VLA-based robotic systems in dynamic environments.

\end{abstract}



\section{Introduction}
\label{sec:intro}

With the development of Vision-Language Models (VLM), Vision-Language-Action (VLA) models~\cite{zitkovich2023rt,kim2024openvla,kim2025fine,black2024pi0,li2023vision} are capable of performing an increasing variety of manipulation tasks from natural-language prompts. These VLA models are typically trained through imitation learning on expert demonstration data, and map high-level task prompts to low-level control actions. However, despite their impressive performance, VLA models often struggle when deployed in dynamic real-world environments, where Out-of-Distribution (OOD) scenarios occur frequently and can significantly degrade performance. 
As a result, VLA models require an accurate runtime monitoring system to determine whether the robot's execution remains consistent with the intended task in OOD situations.

Previous works on runtime monitoring typically formulate failure detection as a robotic state classification problem
~\cite{altan2023clue, sliwowski2024conditionnet, willibald2025multimodal, willibald2025hierarchical, liu2024model}, requiring exhaustive enumeration of abnormal conditions or manually defined preconditions. However, this approach struggles to handle the combinatorial variability of real-world manipulation.
Recent approaches have introduced high-level Vision-Language Models (VLMs), such as in dual-system architectures~\cite{han2024dual, jiang2025galaxea, zhang2024hirt, cui2025openhelix, song2025rationalvla, black2025pi05, nvidia2025gr00t} and Sentinel~\cite{agia2024unpacking}, but these models require multi-step reasoning, resulting in multi-second latency, hindering timely intervention.

To address these limitations, we propose \textbf{R}obot-\textbf{C}onditioned \textbf{N}ormalizing \textbf{F}low (\textbf{RC-NF}), a real-time anomaly detection model that monitors whether the joint distribution of robot states and task-relevant object trajectories is consistent with normal task execution. Specifically, RC-NF is trained solely on successful demonstrations and models robot–object motion through a conditional normalizing flow. At runtime, it computes an anomaly score based on the probability density. If the score exceeds the threshold, behavioral drift is immediately detected. Key to RC-NF is a novel affine coupling layer, the \textbf{R}obot-\textbf{C}onditioned \textbf{P}oint \textbf{Q}uery \textbf{Net}work (\textbf{RCPQNet}), which fuses robot states, task embeddings, and object-centric point-set representations from SAM2~\cite{ravi2024sam} segmentations. In particular, RCPQNet is designed to decouple robot state and object features while preserving their interactive features, providing a structured and expressive representation for manipulation tasks.
%

To evaluate our method, we introduce \textbf{LIBERO-Anomaly-10}, a benchmark for robotic anomaly detection built upon LIBERO-10~\cite{liu2023libero}. 
Unlike prior datasets focused on generalized task failures~\cite{xu2025can, agia2024unpacking,seo2025unisafe,romer2025failure}, LIBERO-Anomaly-10 contains three manipulation-specific anomaly types—\emph{Gripper Open}, \emph{Gripper Slippage}, and \emph{Spatial Misalignment}—each designed to test a different axis of deviation from expert trajectories. 
Compared to current state-of-the-art robotic monitoring model~\cite{xu2025can} and VLMs~\cite{agia2024unpacking,openai2025gpt5,anthropic2025claude45,comanici2025gemini}, our proposed RC-NF achieves state-of-the-art performance across all anomaly types, surpassing the best baseline by approximately 8\% AUC and 10\% AP respectively.

In real-world deployment, RC-NF can detect anomalies within 100 ms. Furthermore, we demonstrate that RC-NF can trigger corrections for both task-level and state-level OOD situations, thereby enhancing the adaptability and robustness of VLA models (e.g., $\pi_0$ ~\cite{black2024pi0}). 
Task-level OOD occurs when the current VLA input task prompt no longer supports the task, requiring task replanning.
State-level OOD refers to situations where the relative motion trajectory between the robot and the object drifts outside the VLA training distribution~\cite{ross2011reduction}. Although the instruction is still valid, recovery can still be achieved by adjusting the trajectory according to the original task.
At the task level, RC-NF serves as a trigger for the high-level module, bridging the low-level and high-level modules. At the state level, RC-NF can trigger the homing procedure to correct the trajectory, enabling real-time adjustments for VLA.

In summary, our main contributions are as follows.
\begin{itemize}
    \item We propose RC-NF, an unsupervised robot-conditioned normalizing flow for robotic anomaly detection, featuring a novel affine coupling layer (RCPQNet) that injects task-aware robot states as conditions into the normalizing flow.
    \item We present LIBERO-Anomaly-10, a new benchmark spanning three diverse manipulation-specific anomaly types, and show that RC-NF achieves state-of-the-art AUC and AP, outperforming prior robotic failure detectors and VLM-based monitors.
    \item We validate RC-NF in real-world robotic manipulation, showing that it detects anomalies within 100 ms and supports both task-level replanning and state-level trajectory rollback, substantially enhancing the robustness and adaptability of VLA models in dynamic environments.
\end{itemize}

\begin{figure*}[t]
  \centering
  \includegraphics[width=\textwidth]{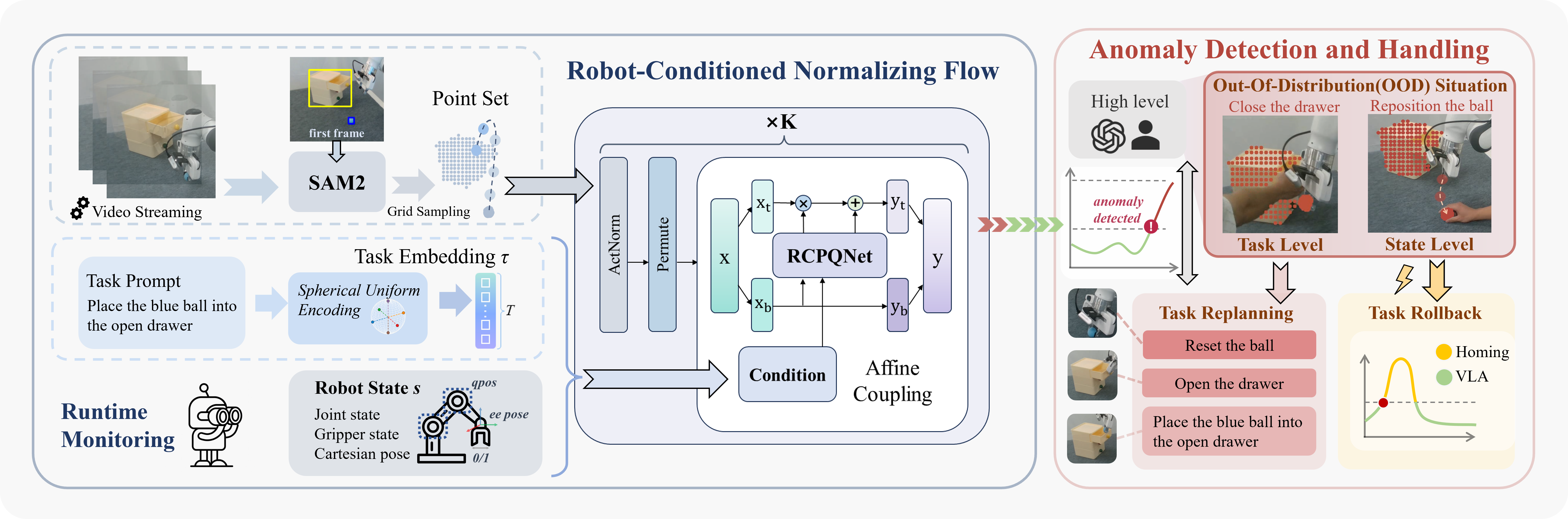}
  \vspace{-4ex}
  \caption{Overview of our framework. Our Robot-Conditioned Normalizing Flow (RC-NF) operates as a real-time runtime monitor for robotic manipulation tasks. (Left) SAM2 extracts object masks from video streaming, which are then grid-sampled into point sets. Task prompts are encoded using spherical uniform encoding, and robot proprioception provides joint, gripper, and pose states. (Center) Our RC-NF leverages these signals as conditions within the affine coupling layers in the proposed RCPQNet (Sec .~\ref {sec:RCPQNet}) to apply $K$ invertible transformations and compute anomaly scores for the current task. 
  (Right) When the anomaly score exceeds the threshold, the Anomaly Detection and Handling module triggers corrective behaviors by task replanning for task-level OOD and task rollback for state-level OOD.}
  \vspace{-3ex}
  \label{fig:method}
\end{figure*}

\section{Related Work}
\label{sec:formatting}
\noindent \textbf{Vision-Language-Action Model.} With the advancement of foundational models, powerful VLA models have emerged~\cite{zitkovich2023rt,kim2024openvla,kim2025fine,li2023vision,black2024pi0}. These imitation learning strategies face challenges during inference, particularly under Out-Of-Distribution (OOD) conditions. For example, task-level OOD occurs when instructions no longer align with the operational environment, while state-level OOD arises when the robotic state drifts from the training distribution~\cite{ross2011reduction,yang2025actor}. Existing methods~\cite{han2024dual,jiang2025galaxea,zhang2024hirt,cui2025openhelix,song2025rationalvla,black2025pi05}that integrate high-level decision-making with low-level execution help address task-level OOD to some extent but are limited by their execution speed and response accuracy. In contrast, our approach can provide feedback on anomalies within 100 ms, effectively correcting low-level execution in response to both task-level and state-level OOD. 

\noindent \textbf{Point Representations for Manipulation.} Points effectively describe the position and motion trajectory of objects. As embodied intelligence progresses, points are vital in both high-level semantic understanding and low-level action execution~\cite{huang2024copa,ji2025robobrain,pan2025omnimanip,huang2024rekep,wen2023any,li2025hamster,wang2023tracking,karaev2024cotracker}. Existing work relies on sparse point trajectories as intermediate representations to guide execution modules. With advancements in open-world object detection~\cite{cheng2024yolo,ren2024dino,fu2025llmdet,xi2025ow} and visual segmentation~\cite{kirillov2023segment,ravi2024sam}, this paper introduces a point set enriched with object semantic shapes, providing a more detailed representation during execution.

\noindent \textbf{Runtime Robot Behavior Monitoring.} Some studies frame anomaly detection as a state classification problem, such as using behavior trees to partition or roll back task steps~\cite{altan2023clue,sliwowski2024conditionnet,willibald2025multimodal,willibald2025hierarchical,liu2024model}. However, these methods depend on explicit task design, which may limit their applicability to more complex tasks. Recent studies have employed VLMs for task monitoring~\cite{agia2024unpacking, wu2025foresight, luo2025visual, wu2025foresight}, but these require chain of thought inferences to generate anomaly assessment, making it difficult to quickly adapt to environmental changes. FailDetect~\cite{xu2025can} proposed an unsupervised flow matching model~\cite{yang2024consistency} that concatenates image features with robot states as its input, but this method leaves potential for improvement in feature selection and processing. To address above, inspired by anomaly detection in pedestrian scenarios~\cite{delic2025sequential, alinezhad2023understanding, hirschorn2023normalizing, wu2025flow, yang2024follow}, RC-NF uses unsupervised training, eliminating the need to enumerate all anomalies. It processes multimodal inputs through decoupled pathways, enabling intuitive anomaly score reflection. As a result, RC-NF can operate as a sub-100 ms fast system, preventing failure accumulation in closed-loop control.
\section{Method}
\subsection{Preliminary}

\noindent{\bf Normalizing Flow (NF)} establishes a bidirectional mapping between the complex distribution \(\mathcal{X}\) and a simple probability distribution \(\mathcal{Z}\) via a sequence of invertible transformations \(f\), enabling explicit probability density estimation:
\begin{equation}
    p_{X}(x)=p_{Z}(f(x))\left|\operatorname{det}\left(\frac{\partial f}{\partial x}\right)\right|.
    \label{NF-2}
\end{equation}

Taking the Glow~\cite{kingma2018glow} model as an example, the entire network consists of \(K\) flow steps \(f\). Each flow step comprises three layers: an ActNorm layer, a permutation layer, and an affine coupling layer. The ActNorm layer applies learnable channel-wise scaling and bias initialized with the first data batch, achieving normalization while preserving invertibility. The permutation transformation employs an invertible \(1\times1\) convolution to reorder feature channels. This formulation allows NF models to compute tractable likelihoods and to model complex distributions while maintaining differentiability and invertibility. The affine coupling layer splits the input into two parts, \( x_0 \) and \( x_1 \), using \( x_0 \) to compute the scale \( \gamma \) and shift \( \beta \) via a neural network. The affine transformation \( x_1 \odot \gamma + \beta \) is then applied to \( x_1 \), leaving \( x_0 \) unchanged. This design ensures invertibility with a tractable Jacobian determinant.

\subsection{Robot-Conditioned Normalizing Flow}
\label{sec:RC-NF}
Building on the foundation of Normalizing Flow, as shown in Fig.~\ref{fig:method}, we propose the Robot-Conditioned Normalizing Flow (RC-NF), which augments Glow with robot- and task-aware conditioning. Specifically, we design a new affine coupling layer—the Robot-Conditioned Point Query Network (RCPQNet, detailed in Sec.~\ref{sec:RCPQNet})—that integrates the robot’s execution state $s$ and target object trajectory $x_{\text{target}}$ to model the conditional probability of successful task execution. This formulation allows RC-NF to detect whether the joint distribution $p_{\mathcal{T}}(x_{\text{target}}, s)$ deviates from the task $\mathcal{T}$ by applying a threshold-based criterion.

Specifically, RC-NF extends a standard NF by conditioning on \( c = (s, \tau) \), where \( s \) denotes the robot state and \( \tau \) is the task embedding. The task embedding \( \tau \) is derived by mapping the current task prompt to a surface vector of a \( T \)-dimensional hypersphere with radius \( \mathcal{R} \). \( T \) is the number of sliding-window steps. These surface vectors, optimized to approximate a uniform distribution on the hypersphere, ensure maximal separation between task embeddings in the latent space, providing an optimal geometric structure for subsequent density estimation. The robot state \( s \) includes the \( T \)-dimensional joint state, gripper state, and Cartesian pose. As illustrated in Fig.~\ref{fig:method}, the video stream captured by the camera is processed by SAM2~\cite{ravi2024sam} to obtain object segmentation masks. Grid sampling is then applied to these masks to extract the shape representation of the manipulated objects. Given this point set \( \mathcal{X} \), RC-NF maps it into a Gaussian latent distribution \( \mathcal{Z} \sim \mathcal{N}(\mu_{\text{task}}, I) \), where the mean \( \mu_{\text{task}} \) is obtained by broadcasting the task embedding \( \tau \) to match the dimensionality of \( \mathcal{Z} \). 

The conditional invertible mapping of RC-NF defines:
\begin{equation}
    z = f_c(x),
    \label{NF-1}
\end{equation}
where \( x \in \mathcal{X} \), \( z \in \mathcal{Z} \), and \( f_c \) is a composition of \( K \) conditional transformations, with intermediate representations \( y_i \), and final output \( z = y_K \):
\begin{equation}
    y_0 = x,\; y_i = f_{i,c}(y_{i-1}),\;z = y_K,\; i = 1, \dots, K
    \label{NF0}.
\end{equation}
Based on the change-of-variables principle, where the Jacobian determinant accounts for the local volume change induced by the transformation, we can compute the conditional likelihood of $x$ given $c$ as follows:
\begin{equation}
    p_{X\mid C}(x \mid c)
    = p_{Z\mid C}\!\left(f_{c}(x)\mid c\right)\;
      \left|\det \frac{\partial f_{c}(x)}{\partial x}\right|
    \label{NF1}.
\end{equation}
Taking the logarithm of Eq.~\ref{NF0} and expanding over $K$ steps:
\begin{equation}
\begin{aligned}
    \log p_{X\mid C}(x \mid c)
    & = \log p_{Z\mid C}(z \mid c) \\
     & + \sum_{i=1}^{K}
        \log \left|\det \frac{\partial f_{i, c}\!\left(y_{i-1}\right)}{\partial y_{i-1}}\right|
    \label{NF2}.
\end{aligned}
\end{equation}
During training, we maximize the log-likelihood in Eq.~\ref{NF2}. For a Gaussian prior over the latent variable $z$, the log-probability term can be expressed as:
\begin{equation}
    \log p_{Z\mid C}(z \mid c) = \text{Const} - \frac{1}{2}\|z - \mu_{\text{task}}\|_2^2.
\end{equation}

The resulting log-probability reflects the likelihood that the current robot-object configuration belongs to the nominal task distribution, with the negative value representing the anomaly score. Each conditional transformation $f_{i, c}$ in RC-NF employs an affine coupling layer parameterized by the Robot-Conditioned Point Query Network (RCPQNet), which fuses task embeddings and robot states to generate context-dependent transformation parameters.

\begin{figure}[t]
  \centering
  \includegraphics[height=0.33\textheight]{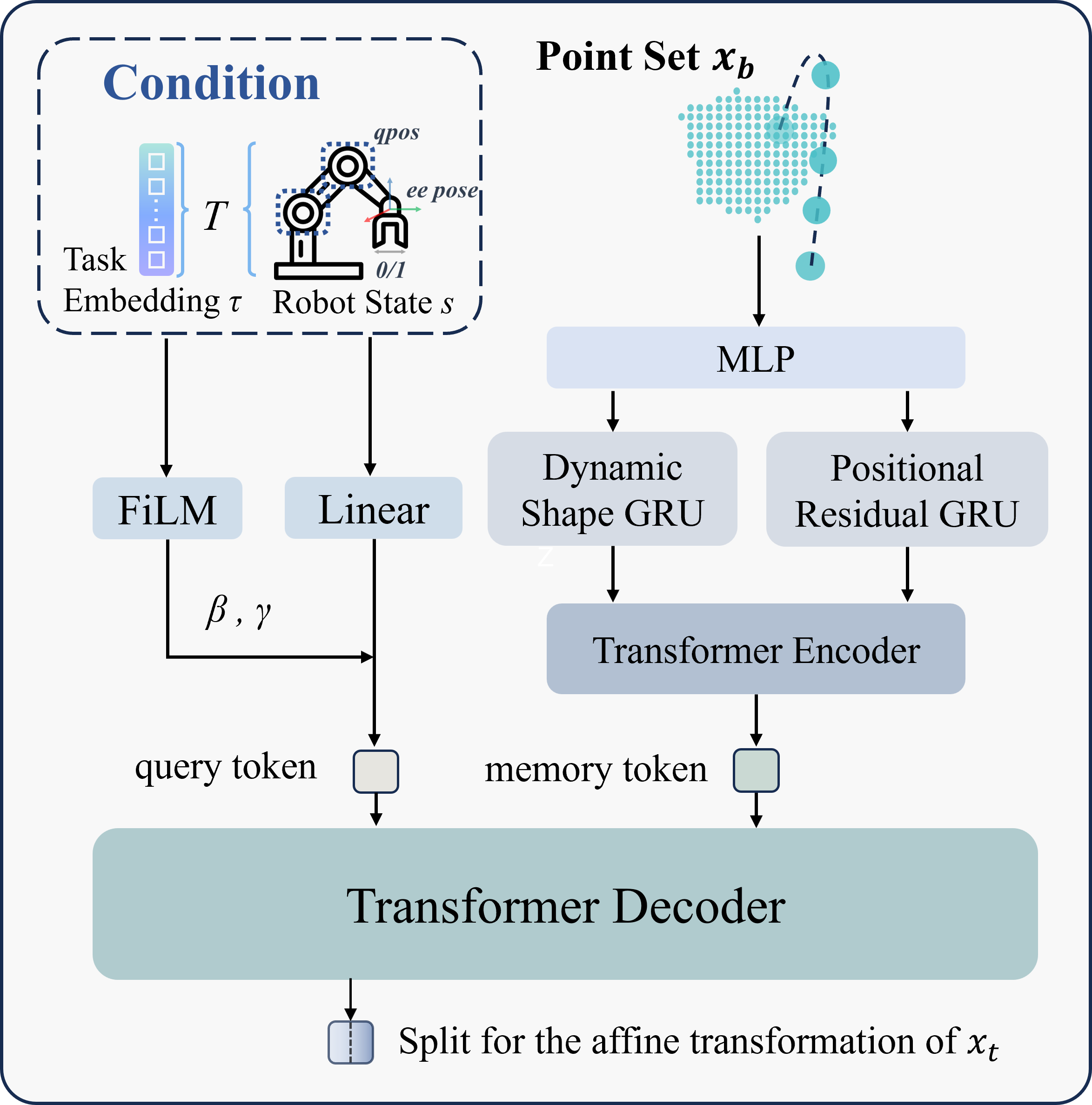}
  \caption{Robot-Conditioned Point Query Network (RCPQNet) is introduced as the affine coupling layer in RC-NF that generates shift and scale parameters.}
  \vspace{-4ex}
  \label{fig:RCPQ}
\end{figure}

\subsection{Robot-Conditioned Point Query Network}
\label{sec:RCPQNet}
As shown in Fig.~\ref{fig:method}, we denote the point set generated by SAM2 as \( x \in \mathbb{R}^{B \times T \times N \times 2} \), where \( B \) represents the batch size, \( T \) is the number of sliding-window steps, \( N \) denotes the number of sampled points per object mask, and the last dimension corresponds to the 2D coordinates of each point. The point set after the ActNorm and Permute layers of the first flow step is also denoted as \( x \), which is used as an example to illustrate the processing flow of the RCPQNet. The point set \( x \) is split into two halves along the temporal dimension: \( x_b = x[:T/2] \) and \( x_t = x[T/2:] \).

The scaling factor \( \gamma \) and shift \( \beta \) are obtained by applying \( \text{RCPQNet}(x_b, c) \), where \( c = (s, \tau) \), as detailed in Sec. ~\ref{sec:RC-NF}:
\[
    \gamma, \, \beta = \text{RCPQNet}(x_b, c).
\]

Based on the above definition, the output of the flow step y, can be expressed as:
\vspace{-1ex}
\[
    y = \begin{bmatrix} y_t \\ y_{\text{b}} \end{bmatrix} = \begin{bmatrix}\gamma \odot x_t + \beta \\ x_b \end{bmatrix}.
\]

As illustrated in Fig.~\ref{fig:RCPQ}, RCPQNet utilizes \textbf{Task-aware Robot-Conditioned Queries} to generate the query token and \textbf{Dual-Branch Point Feature Encoding} to create the memory token. They are then used in cross-attention within the transformer. Hence, the model learns context-aware affine parameters that adjust the flow’s transformation based on the robot’s current state and the performing task.

\noindent \textbf{Task-aware Robot-Conditioned Query.} The robot state is first projected into a latent space through a linear layer and then modulated by the task embedding \(\tau\) using FiLM\cite{perez2018film} to produce task-specific query features. These query tokens encode both the robot’s state context and the high-level goal specified by the task prompt, serving as dynamic keys for retrieving relevant temporal and spatial cues from subsequent feature memories.

\noindent \textbf{Dual-Branch Point Feature Encoding.} To capture complementary spatial-temporal representations, RCPQNet employs Dual-Branch Point Feature Encoding. The \textit{Dynamic Shape branch} performs centering and normalization on each frame to eliminate translation and scale effects during the extraction of shape features. The \textit{Positional Residual branch} focuses on compensating for the information lost during shape normalization of the raw point set. Subsequently, the features are passed through an MLP to increase their dimensionality and then average-pooled to obtain frame-level representations. These representations are sequentially modeled by their respective GRUs to capture temporal dependencies. Finally, they are encoded by a Transformer encoder to produce the memory vectors.

\subsection{ Anomaly Detection and Handling}
\noindent \textbf{Anomaly Detection.} During deployment, RC-NF functions as a plug-and-play and real-time monitoring module that can be seamlessly integrated into existing VLA control loops without architectural modification (see Fig.~\ref{fig:method}). Operating in parallel with the policy network, RC-NF continuously observes both the visual stream and robot-state feedback, estimating the likelihood of the current execution trajectory under the learned nominal task distribution. At each time step, the negative log-likelihood of the observed configuration is used as an anomaly score, where higher values indicate greater deviation from expected task execution.

When the anomaly score exceeds an upper threshold, it indicates both an anomaly and an OOD condition. During RC-NF training, debiasing is applied to ensure the temporal smoothness of anomaly scores in each demonstration. Therefore, we adopt a static threshold that does not vary over time. For each task \( \mathcal{T} \), the upper threshold is estimated using calibration datasets \( S_1 \) and \( S_2 \) derived from successful expert demonstrations\cite{shafer2008tutorial,xu2025can}:

\begin{equation}
\text{Upper}_\mathcal{T} = \mu_\mathcal{T} + Q_{1-\alpha}(D_\mathcal{T}),
\label{threshold}
\end{equation}

where \( \mu_\mathcal{T} \) is the mean anomaly score from \( S_1 \), and \( D_T = \{D_1, \dots, D_{n_2}\} \) denotes the score deviations from \( \mu_\mathcal{T} \) values in \( S_2 \). \( Q_{1-\alpha}(D_\mathcal{T}) \) is the \( (1-\alpha) \)-quantile.

\noindent \textbf{Anomaly Handling.} Once an anomaly is detected, a high-level system determines whether it corresponds to a state-level or task-level OOD case. In practice, the system first performs a lightweight state-level recovery and escalates to task-level replanning if required. On one hand, task-level OOD represents that the environment or context no longer matches the given instruction. For instance, the drawer becomes closed when the prompt specifies \textit{place the blue ball into the open drawer}. In this case, RC-NF alerts the high-level controller (e.g, a human operator or LLM-based planner) to perform task replanning, producing an updated sequence of sub-tasks or object interactions consistent with the new environmental state. On the other hand, state-level OOD indicates that the task remains valid, but the robot’s physical configuration has drifted from the nominal distribution. For example, during the execution of the task \textit{place the blue ball into the open drawer}, if the ball falls from the gripper onto the table (in this experiment, we manually reposition the ball from the gripper onto the table), RC-NF detects an increase in the anomaly score. Then a homing procedure, designed to return the robotic arm to its initial state, is activated for task rollback. It locally adjusts the trajectory or control parameters until the anomaly score falls below the threshold. Once the environment and robot state return to the familiar data distribution, execution control is seamlessly handed back to the VLA for normal operation. 

\begin{table*}[!t]
\vspace{-3ex}
\caption{Quantitative comparison of anomaly detection performance on the LIBERO-Anomaly-10 benchmark.}
\vspace{-1ex}
\label{tab:sim1}
\centering
\renewcommand{\arraystretch}{1.2}
\setlength{\tabcolsep}{5pt}
\begin{tabular}{lcc cc cc cc}
\toprule
\textbf{Method} &
\multicolumn{2}{c}{\textbf{Gripper Open}} &
\multicolumn{2}{c}{\textbf{Gripper Slippage}} &
\multicolumn{2}{c}{\textbf{Spatial Misalignment}} &
\multicolumn{2}{c}{\textbf{Average}} \\
\cmidrule(lr){2-3}
\cmidrule(lr){4-5}
\cmidrule(lr){6-7}
\cmidrule(lr){8-9}
 & \textbf{AUC} & \textbf{AP} & \textbf{AUC} & \textbf{AP} & \textbf{AUC} & \textbf{AP} & \textbf{AUC} & \textbf{AP} \\
\midrule
GPT-5~\cite{openai2025gpt5}            & 0.9137 & 0.9642 & 0.8941 & 0.8720 & 0.4904 & 0.4015 & 0.8500 & 0.8507 \\
Gemini 2.5 Pro~\cite{comanici2025gemini}  & 0.8644 & 0.9333 & 0.8633 & 0.8505 & 0.5167 & 0.4271 & 0.8186 & 0.8313 \\
Claude 4.5~\cite{anthropic2025claude45}      & 0.8754 & 0.9401 & 0.8551 & 0.8285 & 0.5292 & 0.4290 & 0.8214 & 0.8249 \\
FailDetect~\cite{xu2025can}                 & 0.7883 & 0.9032 & 0.6665 & 0.6932 & 0.6557 & 0.5820 & 0.7181 & 0.7700 \\
\textbf{RC-NF(Ours)}              & \textbf{0.9312} & \textbf{0.9781} & \textbf{0.9195} & \textbf{0.9180} & \textbf{0.9676} & \textbf{0.9585} & \textbf{0.9309} & \textbf{0.9494} \\
\bottomrule
\end{tabular}
\vspace{-3ex}
\end{table*}

\section{Experiment}
We primarily conduct quantitative analysis of RC-NF's anomaly detection capability through simulation experiments, and demonstrate the significance of the monitoring mechanism through real-world robotic tasks.
\subsection{Experimental Setup}
\label{sec:exp_setup}
\noindent{\bf Datasets.} We train RC-NF on the original LIBERO-10 dataset, which efficiently encompasses all the scenarios present in the LIBERO dataset\cite{liu2023libero}, with 50 demonstrations per task. We evaluate it on our proposed LIBERO-Anomaly-10 benchmark. It introduces three distinct anomaly types, illustrated in Fig.~\ref{fig:main}:
(1) \textbf{\textit{Gripper Open}}: the gripper remains open at $t_{\text{anomaly}}$, when it should grasp the object, leaving the object untouched and unperturbed. This case tests a situation where the object position remains normal, but the robot state does not align with it. (2) \textbf{\textit{Gripper Slippage}}: at $t_{\text{anomaly}}$, when the gripper is holding the object, the gripper friction is set to zero, causing the object to slip and produce abnormal trajectory fluctuations.
(3) \textbf{\textit{Spatial Misalignment}}: the robot moves toward the wrong compartment (left, front or right) instead of the intended \textbf{back} at $t_{\text{anomaly}}$, testing semantic–spatial misalignment between the task prompt and motion direction.

During annotation, the period before $t_{\text{anomaly}}$ is labeled as normal and after $t_{\text{anomaly}}$ as abnormal. \textit{Gripper Open} and \textit{Gripper Slippage} are generated for all ten tasks in LIBERO-10, whereas \textit{Spatial Misalignment} corresponds to tasks involving three different incorrect compartment placements (front, left, and right), rather than the demo task \textit{Pick up the book and place it in the back compartment of the caddy}.

\noindent{\bf Baselines and Evaluation Metrics.} We compare RC-NF against two categories of methods. The first category is VLM-based monitoring method, represented by Sentinel~\cite{agia2024unpacking}, where we employ the most advanced VLMs, GPT-5~\cite{openai2025gpt5}, Gemini 2.5 Pro~\cite{comanici2025gemini}, and Claude 4.5~\cite{anthropic2025claude45}. They are evaluated under conditions involving parallel invocation and a sampling frequency of 1 Hz. The second category is also flow-based, represented by FailDetect~\cite{xu2025can}, a flow-matching model that jointly encodes image and robot-state features for unsupervised failure detection.

We evaluate the model using Area Under the ROC Curve (AUC) and Average Precision (AP). AUC assesses the model's overall discriminatory ability, while AP focuses on precision-recall performance, particularly for imbalanced tasks like anomaly detection. These threshold-independent metrics offer a comprehensive evaluation of both discriminatory power and precision-recall trade-offs.

\noindent{\bf Real-World Setup.} 
Experiments are conducted on a Franka Research 3 robotic arm equipped with both a wrist-mounted and a third-person camera, both of which are Intel RealSense D435 depth cameras. VLA model $\pi_0$~\cite{black2024pi0} is adopted as the imitation policy baseline. The training demonstrations for both VLA and RC-NF are consistent. VLA uses both camera views during training, while RC-NF uses only the third-person camera for monitoring. The threshold parameters of real-world experiments in Eq.~\ref{threshold} are set with \( \alpha = 0.05 \), following the design of FailDetect~\cite{xu2025can}.

\noindent{\bf Implementation Details.} As shown in Fig.~\ref{fig:method}, before SAM2 processes the video stream, a bounding box prompt is required for the first frame. In the simulation environment, computer graphics techniques are used to ensure the reproducibility of the point set. In the real-world setup, the bounding box is obtained using the Gemini 2.5 Pro.

The parameters of FailDetect~\cite{xu2025can} follow its open-source code exactly. The prompts for the VLMs are derived from Sentinel~\cite{agia2024unpacking}, originally designed for task progress assessment, and are used in this paper as anomaly scores with the following anchors: 0–3 (normal), 4–6 (slightly abnormal), and 7–10 (strongly abnormal). For RC-NF, the number of flow steps is \( K = 12 \) and the model is trained for 100 epochs with debiasing applied for data balancing. More details can be found in the Supplementary Materials.

\begin{figure}[!t]
  \centering
  \includegraphics[width=\columnwidth]{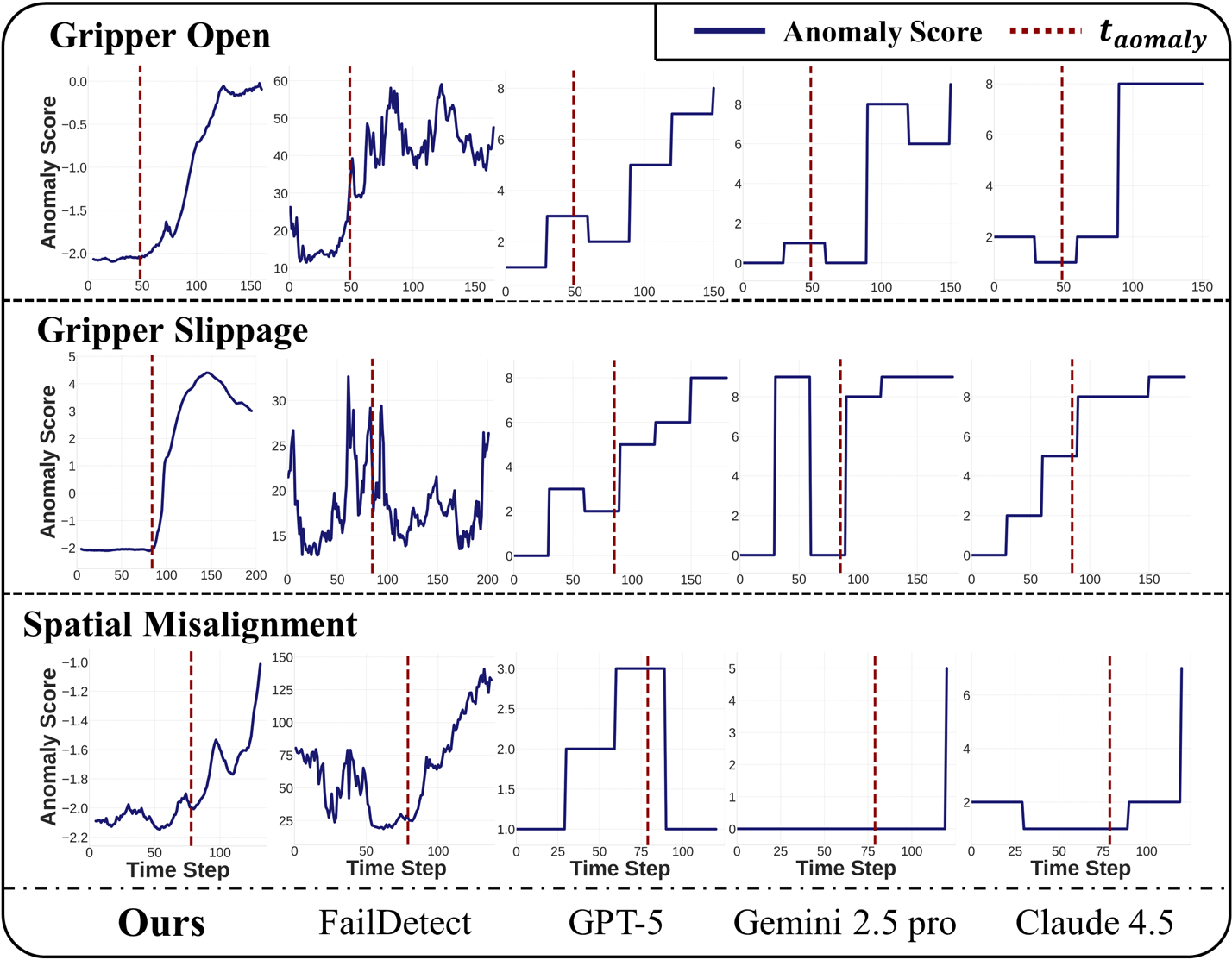}
  \vspace{-3ex}
  \caption{Visualization of the anomaly scores for the task \textit{Pick up the book and place it in the back compartment of the caddy}. The x-axis denotes the time steps, and the y-axis indicates the anomaly score. The red dashed line marks $t_{\text{anomaly}}$.}
  \vspace{-4ex}
  \label{fig:exp_sim1}
\end{figure}

\subsection{Performance Comparison}
\label{sec:sim_exp}
Table~\ref{tab:sim1} presents a comparison of anomaly detection performance on the LIBERO-Anomaly-10 benchmark.
Across all anomaly types, RC-NF consistently outperforms baselines across all metrics, surpassing
the best prior method by about \textbf{8\% in AUC} and \textbf{10.0\% in AP} on average. 

VLMs achieve moderate success in \textit{Gripper Open} and \textit{Gripper Slippage} but degrade to near-random performance (AUC $\approx$ 0.5, AP $\approx$ 0.4) on \textit{Spatial Misalignment}, highlighting their difficulty in aligning visual semantics with spatial reasoning. In contrast, RC-NF can precisely compute the degree of deviation from the normal trajectory for the task in real-time, replacing semantic understanding with the spatial trajectories of the object and robot arm.

Compared to FailDetect, RC-NF offers significant advantages in feature selection and feature processing for robotic anomaly monitoring. In terms of feature selection, RC-NF operates on a refined point-set representation derived from segmentation masks, which offers greater robustness to noise than raw image features used in FailDetect. Regarding feature processing, RC-NF treats the robot state features as query tokens and the object point features as memory tokens, enabling an efficient and decoupled design that encourages multimodal information interaction while avoiding feature interference.
By contrast, FailDetect concatenates robot and image features as flow-matching inputs, leading to entangled representations and feature imbalance.
 
Fig.~\ref{fig:exp_sim1} visualizes the anomaly detection curves for the task \textit{Pick up the book and place it in the back compartment of the caddy}. RC-NF maintains a stable score of -2 before \( t_{\text{anomaly}} \) for each task, indicating normal behavior. Once an anomaly occurs after \( t_{\text{anomaly}} \), as indicated by the red dashed line, it quickly reacts by increasing the anomaly score. Compared to VLM-based monitoring methods, RC-NF can calculate the anomaly score in real-time without the need for prolonged inference or reasoning. In contrast to FailDetect, RC-NF produces scores that are superior in terms of smoothness, stability, and accuracy, which can be attributed to its structured feature decoupling and robot-conditioned flow formulation.

\begin{table*}[!t]
\vspace{-3ex}
\caption{Ablation study of the proposed Robot-Conditioned Point Query Network (RCPQNet). RC-NF (Ours) denotes the full model.}
\vspace{-1ex}
\label{tab:sim2}
\centering
\renewcommand{\arraystretch}{1.2}
\setlength{\tabcolsep}{5pt}
\begin{tabular}{r l cc cc cc cc}
\toprule
Row & \textbf{Method} &
\multicolumn{2}{c}{\textbf{Gripper Open}} &
\multicolumn{2}{c}{\textbf{Gripper Slippage}} &
\multicolumn{2}{c}{\textbf{Spatial Misalignment}} &
\multicolumn{2}{c}{\textbf{Average}} \\
\cmidrule(lr){3-4}
\cmidrule(lr){5-6}
\cmidrule(lr){7-8}
\cmidrule(lr){9-10}
 & & \textbf{AUC} & \textbf{AP} & \textbf{AUC} & \textbf{AP} & \textbf{AUC} & \textbf{AP} & \textbf{AUC} & \textbf{AP} \\
\midrule
\setcounter{enumi}{0} 
1 & \textbf{RC-NF(Ours)}      & \textbf{0.9312} & \textbf{0.9781} & \textbf{0.9195} & \textbf{0.9180} & \textbf{0.9676} & \textbf{0.9585} & \textbf{0.9309} & \textbf{0.9494} \\
2 & \hspace{1em}w/o Task Embedding        & 0.8769 & 0.9603 & 0.8668 & 0.8680 & 0.8139 & 0.8118 & 0.8643 & 0.9008 \\
3 & \hspace{1em}w/o Robot State           & 0.6327 & 0.8621 & 0.7443 & 0.8116 & 0.8929 & 0.8617 & 0.7152 & 0.8401 \\
\hline
4 & \hspace{1em}w/o Pos. Residual branch     & 0.9045 & 0.9712 & 0.8971 & 0.9085 & 0.8543 & 0.8072 & 0.8947 & 0.9225 \\
5 & \hspace{1em}w/o Dyn. Shape branch        & 0.7666 & 0.9234 & 0.7763 & 0.8108 & 0.1022 & 0.2755 & 0.6841 & 0.7899 \\
\bottomrule
\end{tabular}
\vspace{-2ex}

\end{table*}


\begin{figure}[t]
  \centering
  \includegraphics[width=\columnwidth]{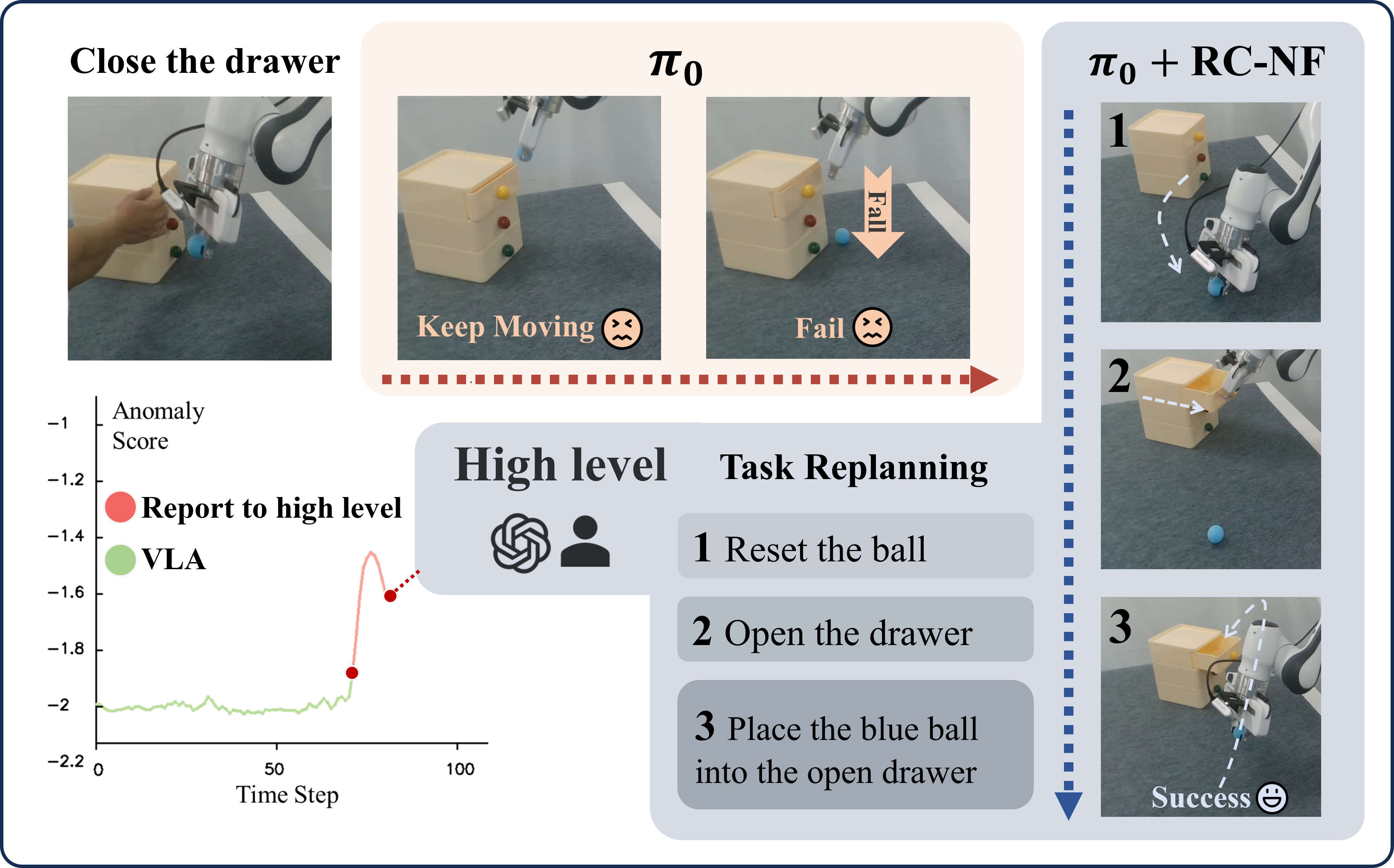}
  \vspace{-3ex}
  \caption{A real-world comparison of $\pi_0$ and $\pi_0$ + RC-NF (ours) during the task \textit{placing the ball into the open drawer}, when the drawer closes midway.}  
  \vspace{-4ex}
  \label{fig:exp_real0}
\end{figure}

\subsection{Ablation Study}
\label{sec:ablation}
To investigate the effectiveness of the proposed Robot-Conditioned Point Query Network (RCPQNet), as shown in Tab.~\ref{tab:sim2}, we perform ablations on the two principal components introduced in Sec. \ref{sec:RCPQNet}.

\noindent \textbf{Effect of Task-Aware Robot-Conditioned Query.} To assess the role of task modulation and robot awareness, we remove either the task embedding or the robot-state features. When the task embedding is removed (Row 2), the point set \( \mathcal{X} \) is projected into the latent space \( \mathcal{Z} \sim \mathcal{N}(0, I) \). As a result, model can only detect dataset-level OOD samples but fails to distinguish task-specific anomalies. Performance on \textit{Spatial Misalignment} drops nearly threefold compared to the counterparts, confirming that task embeddings are crucial for encoding spatial intent and disambiguating similar object trajectories under different instructions. In addition, removing robot-state features (Row 3) eliminates contextual grounding of object motion. For the \textit{Gripper Open anomaly}, since the gripper remaining open does not directly displace the object, the anomaly manifests in the relative motion between the robot and the object, leading to the greatest drop in AUC. In contrast, for the \textit{Spatial Misalignment} anomaly, the anomaly can be detected through the object's positional deviation, leading to the least decline. These results highlight that robot-conditioned query enables RC-NF to reason jointly about robot–object motion, resulting in more precise anomaly localization.

\noindent \textbf{Effect of Dual-Branch Point Feature Encoding.} We further examine the contribution of the two encoding branches within RCPQNet: the \textit{Positional Residual branch} and the \textit{Dynamic Shape branch}. Removing the positional branch (Row 4) results in a moderate drop, with an average AUC of approximately 0.89, suggesting that absolute spatial information complements dynamic-shape cues by preserving average shifts within robot-object motion features. The \textit{Dynamic Shape branch} treats the point sets of all objects as a unified whole, representing the relative movements between target objects as shape variations of the entire set. Removing this branch (Row 5) leads to the greatest degradation (AUC $\approx$ 0.68 on average), underscoring its importance in Dual-Branch Point Feature Encoding. The sharp decline observed under \textit{Spatial Misalignment} further indicates that temporal shape evolution provides stronger evidence of abnormality than inter-frame average positional shifts. 

\begin{figure*}[t]
  \centering
  \includegraphics[width=\textwidth]{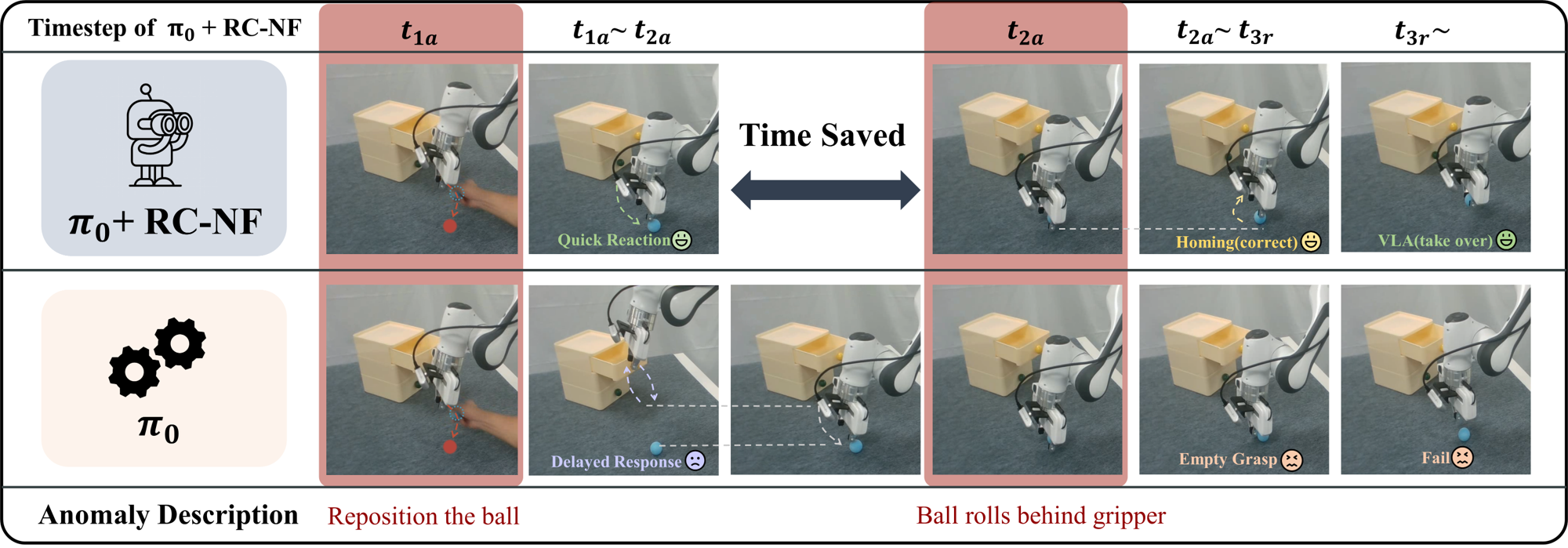}
  \vspace{-2ex}
  \caption{Comparison of \( \pi_0 \) + RC-NF (Ours) and \( \pi_0 \) model in handling two consecutive task-level OOD events during real-world deployment. \textbf{$t_{1a}$} represents the moment when the ball is repositioned on the table. \textbf{$t_{2a}$} marks when the ball rolls behind the gripper. \textbf{$t_{3r}$} indicates when control is returned to the VLA after homing. The timeline of \( \pi_0 \) + RC-NF (Ours) is shown in Fig.~\ref{fig:exp_real2}. 
}
  \vspace{-3ex}
  \label{fig:exp_real1}
\end{figure*}

\begin{figure}[t]
  \centering
  \includegraphics[width=\columnwidth]{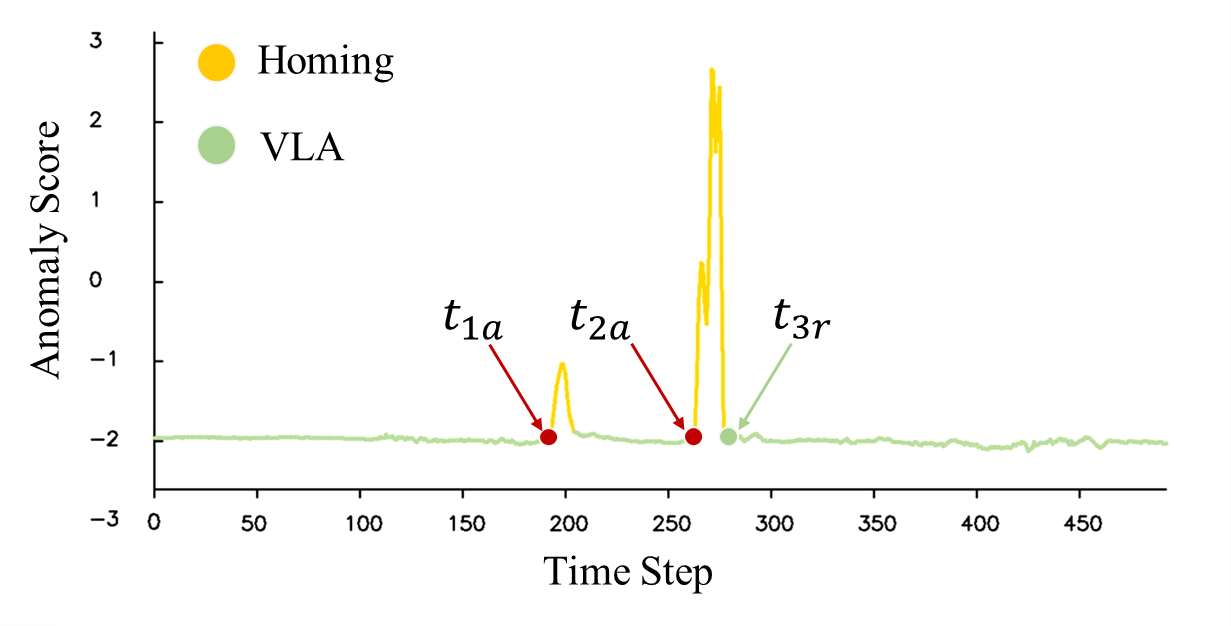}
  \vspace{-4ex}
  \caption{ Anomaly score plot for a state-level OOD sequence where the ball is repositioned at $t_{1a}$ and rolls behind the gripper at $t_{2a}$.
  The yellow curve indicates the detection of an anomaly, with the trajectory corrected by the homing procedure. The green curve represents the operation of the VLA model $\pi_0$.}
  \vspace{-3ex}
  \label{fig:exp_real2}
\end{figure}

\subsection{Real-World Evaluation}

We further conduct real-world evaluations to demonstrate how RC-NF enhances the adaptability and robustness of Vision-Language-Action (VLA) policies. The study focuses on the representative scenario \textit{placing the blue ball into the open drawer} and associated derivative tasks using the \( \pi_0 \) model as the baseline. In real-world deployment, RC-NF can detect anomalies within 100 ms on a consumer-grade GPU such as Nvidia GeForce RTX 3090. Related videos can be found in the Supplementary Materials.

\noindent \textbf{Task-Level Anomaly Handling.} As shown in Fig.~\ref{fig:exp_real0}, during the execution of the task, if the drawer closes unexpectedly,  \( \pi_0 \) policy continues execution and fails. With our RC-NF, the anomaly score computed from the visual stream and robot state increases sharply beyond the threshold defined in Eq.~\ref{threshold}. \( \pi_0 \)+RC-NF manages to pause the motion and signals the high-level controller (human or LLM planner) to initiate task replanning, i.e., reset the ball and open the drawer first, then execute the task normally, leading to task success.

\noindent \textbf{State-Level Anomaly Handling.} As shown in Fig. \ref{fig:exp_real1} and \ref{fig:exp_real2}, we next examine state-level disturbances that preserve task intent but perturb execution dynamics. At \textbf{$t_{1a}$}, when the ball returns to the table, RC-NF quickly signals an anomaly and activates the homing procedure to re-align the trajectory. Nevertheless, $\pi_0$ alone experiences a delay, which continues moving to the drawer with empty gripper. In addition,  when the ball rolls behind the gripper at \textbf{$t_{2a}$}, RC-NF detects the anomaly immediately and adjusts the trajectory towards the new position of the ball. Once the anomaly score drops to normal at \( t_{3r} \), control is seamlessly returned to VLA for regular operation. Without RC-NF, $\pi_0$ continues to move incorrectly, causing the gripper to grasp air and lift.

The real-world evaluation demonstrates that RC-NF transfers effectively from simulation to physical settings. It provides low-latency feedback for both task-level and state-level OOD anomalies, enabling safe and adaptive execution without altering the underlying VLA architecture. The plug-and-play design allows direct integration with existing imitation-learning policies, enhancing robustness and situational awareness in unstructured environments.

\section{Conclusion}
We introduced the Robot-Conditioned Normalizing Flow model, designed to enhance the adaptability of VLA models in dynamic real-world environments. By utilizing \textit{Task-aware Robot-Conditioned Query} and \textit{Dual-Branch Point Feature Encoding}, RC-NF enables real-time robotic anomaly detection, improving the robot's ability to handle Out-of-distribution scenarios. Our quantitative experiments on the LIBERO-Anomaly-10 benchmark show that RC-NF outperforms existing methods for monitoring, achieving state-of-the-art AUC and AP scores. These results underscore RC-NF's effectiveness in monitoring robotic tasks, detecting deviations in task execution and object motion. RC-NF acts as a real-time monitor that produces a general-purpose OOD signal meaningful across different granularities. Real-world evaluations further confirm RC-NF’s effectiveness, highlighting its fast response (under 100 ms) and seamless integration with the VLA policy (e.g. \( \pi_0 \)).
\section*{Acknowledgments}
This work was supported by the Science and Technology Commission of Shanghai Municipality (No. 24511103100) and the National Natural Science Foundation of China (NSFC) under Grant No. 62521004. It was also supported by the Ministry of Education (MOE), Singapore, under its Academic Research Fund (AcRF) Tier 2 (Proposal ID: T2EP20125-0048). Any opinions, findings, and conclusions or recommendations expressed in this material are those of the authors and do not necessarily reflect the views of the Ministry of Education, Singapore.
{
    \small
    \bibliographystyle{ieeenat_fullname}
    \bibliography{main}
}

\clearpage
\setcounter{page}{1}
\setcounter{section}{0}
\maketitlesupplementary

\section{Experimental Demonstration Videos}
\textbf{Simulated\_Environment\_Results.mp4} presents three anomalies of LIBERO-Anomaly-10: Gripper Open, Gripper Slippage, and Spatial Misalignment, along with the corresponding anomaly scores inferred by RC-NF.\\ \textbf{Real\_World\_Experiments\_Results.mp4} presents real-world experiments, exhibiting how the RC-NF monitoring mechanism enhances the robustness and adaptability of the model \( \pi_0 \), in task-level and state-level OOD situations.

\section{Further details on Task Embedding}
\begin{figure*}[h]
    \centering
    \includegraphics[width=\linewidth]{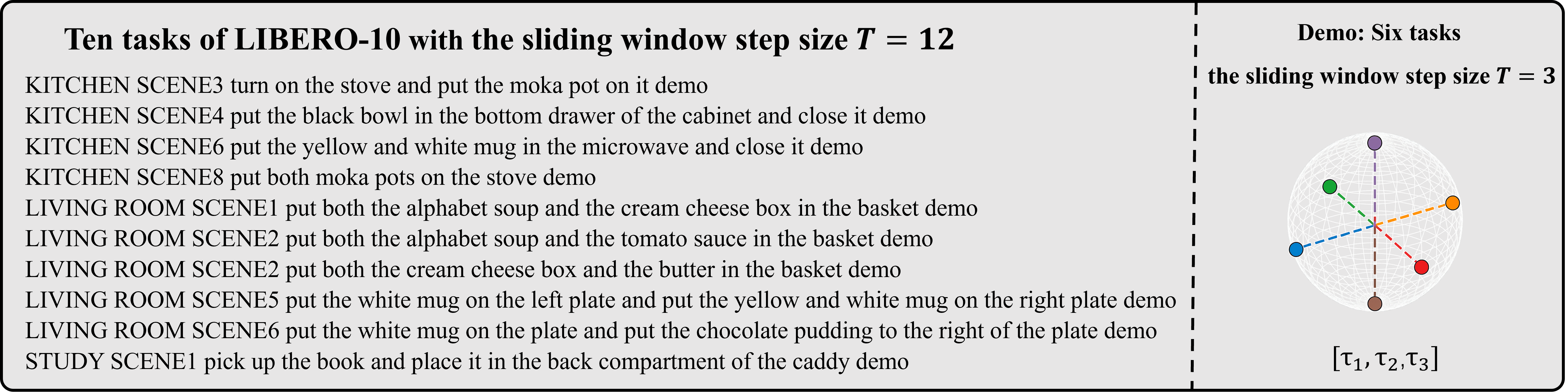}
    \caption{Diagram illustrating Spherical Uniform Encoding. The left side displays the ten tasks from the training set along with their descriptions, while the right side provides an example demonstration.}
    \label{fig:suppl_task_emb}
\end{figure*}

\noindent \textbf{Implementation Details of Task Embedding.} This section explains the implementation of task embedding $\tau$ from Sec. \ref{sec:RC-NF}. Fig. \ref{fig:suppl_task_emb} (left) shows the ten tasks in the LIBERO-10. In deployment, the first step is to obtain 10 vectors uniformly distributed on a hypersphere with radius \( R = 5 \) and dimension \( T = 12 \) using an optimization algorithm. Then, these 10 task instructions are mapped one-to-one to the spherical vectors, resulting in \([\tau_1, \tau_2, \dots, \tau_{12}]\). Fig. \ref{fig:suppl_task_emb} (right) provides detailed examples to clarify the mapping process. When \( T = 3 \) and the number of tasks is 6, this results in 6 spherical vectors on a 3-dimensional sphere, intersecting the \( x \), \( y \), and \( z \)-axes, as represented by \([\tau_1, \tau_2, \tau_3]\).

\begin{figure*}[h]
    \centering
    \includegraphics[width=0.8\linewidth]{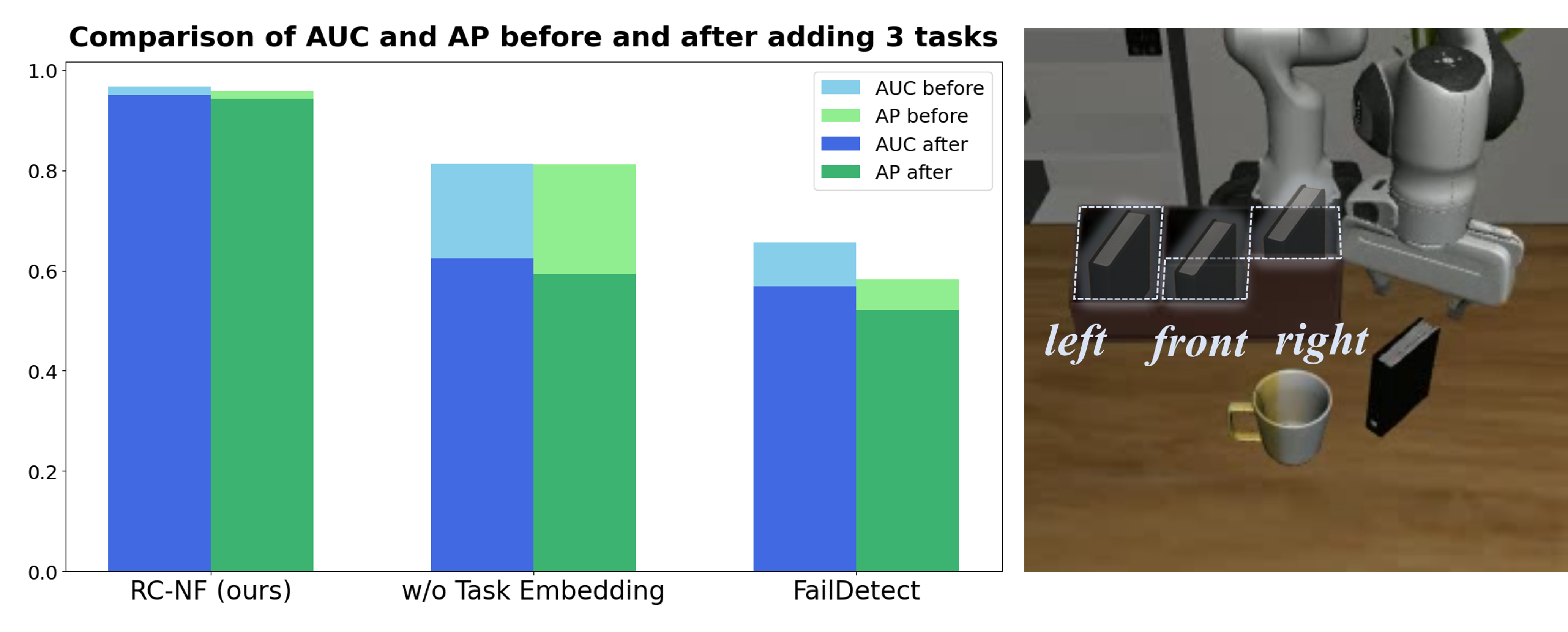}
    \caption{Comparison of performance metrics before and after incorporating three tasks with spatial misalignment (left, front and right) into the training set.}
    \label{fig:suppl_sim_te}
\end{figure*}

\noindent \textbf{Role of Task Embedding.} As discussed in Sec. \ref{sec:ablation}, removing the task embedding limits the model to detecting only dataset-level OOD samples, preventing it from identifying task-specific anomalies. To illustrate this, consider the Spatial Misalignment setup: during training, the book is placed in the back compartment, while in the test scenario, the robotic arm places the book in the left, front, or right compartments, referred to as the 3 tasks (see Fig. \ref{fig:suppl_sim_te}(right)). Tables~\ref{tab:sim1} and~\ref{tab:sim2} show that without these tasks in the training set, the action trajectories are out-of-distribution, and Spatial Misalignment is classified as a dataset-level OOD anomaly. When these 3 tasks are included in the training set, the action trajectories become in-domain, and detecting Spatial Misalignment is recognized as a task-specific anomaly. Fig. \ref{fig:suppl_sim_te} compares the metrics for this scenario. After adding the 3 tasks, the performance of the RC-NF model shows almost no decrease. However, without the task embedding, the RC-NF monitoring performance drops to around 0.6, emphasizing the importance of task embedding in distinguishing tasks within the training set. In contrast, FailDetect’s performance drops to 0.5, highlighting its inability to differentiate tasks within the same training set.

\begin{algorithm}[h]
\caption{Training Loop with BalancedHardSampler}
\label{al:Training Loop}
\begin{algorithmic}[1]
\For{epoch $\leftarrow$ startEpoch to numEpochs}
    \State model.train()
    \For{data in trainLoader}
        \State \texttt{...  // process data, compute loss, update model}
    \EndFor
    \If{epoch = NextStageEpoch}
        \State model.eval()
        \For{data in trainLoader}
            \State update\_weights(model(data))
        \EndFor
        \State sampler $\leftarrow$ BalancedHardSampler(weights)
    \EndIf

    \If{epoch $\geq$ NextStageEpoch}
        \State update\_train\_loader(sampler)
    \EndIf
\EndFor
\end{algorithmic}
\end{algorithm}

\section{Training Data Debiasing}
The Training Data Debiasing Procedure aims to address the issue of imbalanced sample distribution during robotic manipulation. Specifically, in the initial stages of a robotic arm's motion, the motion trajectories tend to be more consistent, whereas near the object grasping phase, the trajectories exhibit greater variation. This results in a higher frequency of similar trajectories from the initial movement phase, causing an uneven sample distribution. To mitigate this, we design the \textbf{BalancedHardSampler} class, designed to equalize the distribution of these samples, ensuring a more balanced and robust training process.

Algorithm \ref{al:Training Loop} illustrates the training procedure, which is divided into two stages. In the first stage, standard sampling is used to train RC-NF. Upon reaching the NextStageEpoch, inference results are collected for sample evaluation. The BalancedHardSampler is then initialized to reduce redundant samples while increasing the diversity of underrepresented samples, ensuring a more balanced sample distribution throughout the demonstration. After the NextStageEpoch, the BalancedHardSampler is continuously employed for sample collection during the remaining stages.

\section{Implementation Details for Simulations}

As mentioned in the Implementation Details in Sec. \ref{sec:exp_setup}, computer graphics techniques are used in the simulation environment to generate the first frame's bounding box for SAM2. This method of obtaining the bounding box is stable, low-cost, and ensures reproducibility. The following section provides a detailed explanation of the process.

\noindent \textbf{Generating Sample Points in Geometry.} For each object, we generate a grid of sample points within the geometry. Assuming the geometry size is \( \text{geom\_size} = [W, H, D] \), where \( W \), \( H \), and \( D \) represent the width, height, and depth of the geometry, respectively, the offset for each sample in a 5x5x5 grid is computed as:

\[
\text{local\_offset} = \begin{bmatrix} dx \cdot W \\ dy \cdot H \\ dz \cdot D \end{bmatrix}.
\]

Here, \( dx, dy, dz \) are the offsets in the grid along the width, height, and depth axes of the geometry.

These local offsets are then transformed into world coordinates using the geometry’s position in the world \( \text{geom\_pos} \) and its rotation matrix \( R \):

\[
P_{\text{world}} = \text{geom\_pos} + R \cdot \text{local\_offset}.
\]

In this equation, \( \text{geom\_pos} \) is the position of the geometry in the world frame (a 3D vector representing the center of the geometry), and \( R \) is the rotation matrix that transforms the local offset into the world frame.

\noindent \textbf{Projection of 3D Points to 2D.} After obtaining the world coordinates of all sample points \( P_{\text{world}} \), we project them onto the 2D image plane using the camera's intrinsic matrix \( K \) and the extrinsic matrix \( \text{Extrinsic} \):

\[
P_{\text{image}} = K \cdot (\text{Extrinsic} \cdot P_{\text{world}}).
\]

In this equation, \( \text{Extrinsic} \) represents the transformation from the world coordinates to the camera coordinates (a combination of rotation and translation), and \( P_{\text{image}} = [x, y]^T \) represents the 2D coordinates of the projected point on the image plane.

\noindent \textbf{Filtering Invalid Projections.} We filter out invalid projections, such as those behind the camera or outside the image boundaries. For each projected point \( P_{\text{image}} = [x, y] \):

\[
0 \leq x < \text{image\_width}, \quad 0 \leq y < \text{image\_height}.
\]

\noindent \textbf{Bounding Box Calculation.} For the valid projected points, we calculate the bounding box by finding the minimum and maximum \( x \) and \( y \) coordinates:

\[
x_{\text{min}} = \min(x), \quad x_{\text{max}} = \max(x).
\]
\[
y_{\text{min}} = \min(y), \quad y_{\text{max}} = \max(y).
\]

The bounding box is then defined as:

\[
\text{bbox} = [x_{\text{min}}, y_{\text{min}}, x_{\text{max}}, y_{\text{max}}].
\]

\section{Implementation Details in the Real World}

\begin{figure*}[h]
    \centering
    \includegraphics[width=0.8\linewidth]{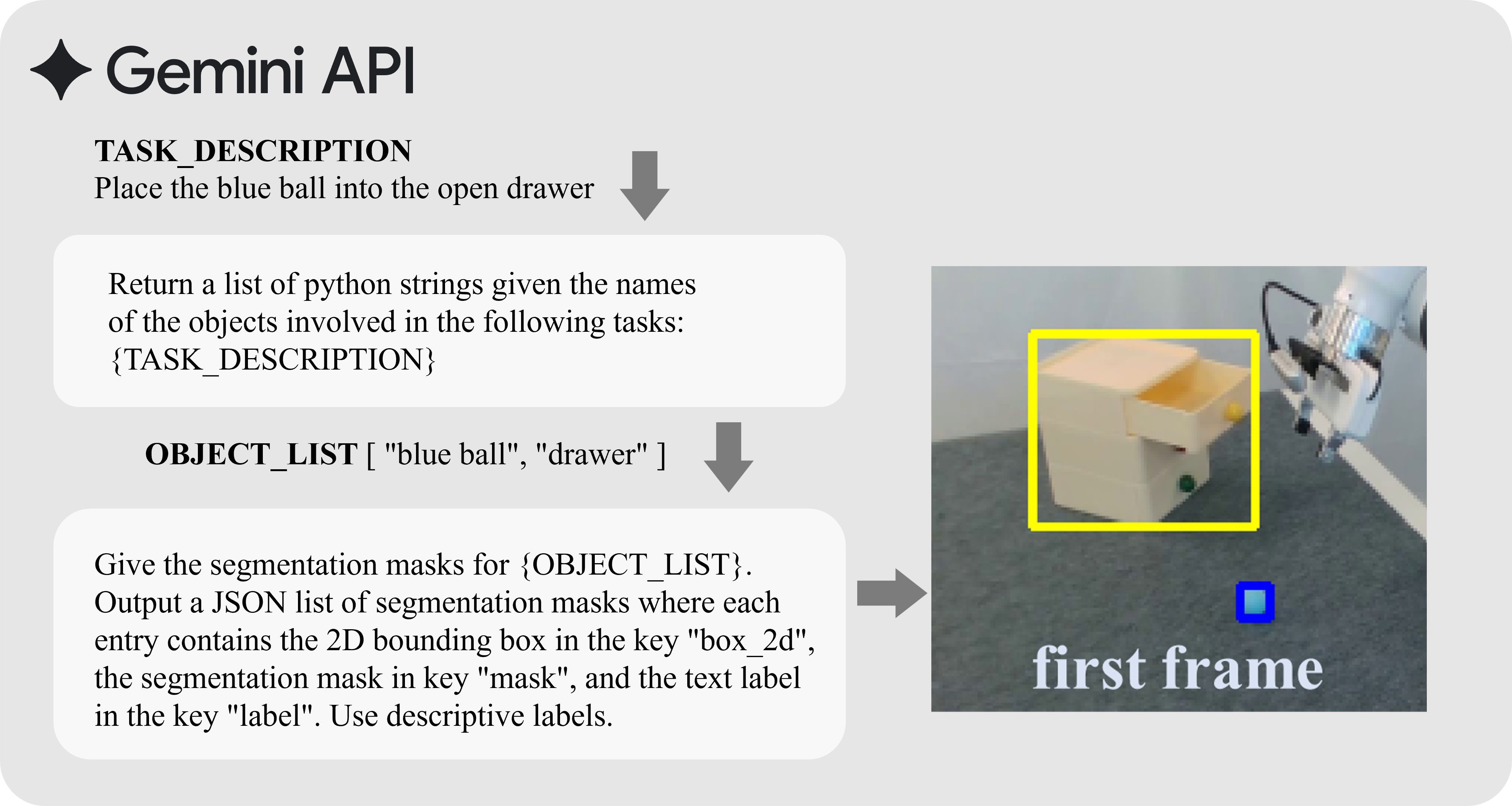}
    \caption{Prompts for generating bounding boxes with Gemini 2.5 Pro}
    \label{fig:suppl_prompt_bbox}
\end{figure*}

\noindent \textbf{Bounding Box Generation.} In this paper, we utilize the multimodal large model Gemini 2.5 Pro\cite{comanici2025gemini}, which can understand the manipulated object from task instructions and generate a zero-shot bounding box for SAM2. As illustrated in Fig. \ref{fig:suppl_prompt_bbox}, the process begins by extracting the OBJECT\_LIST (e.g., \textit{blue ball}, \textit{drawer}) from the TASK\_DESCRIPTION, such as \textit{Place the blue ball into the open drawer}. The image and identified objects are subsequently passed to Gemini, which generates a JSON list.

\noindent \textbf{Description of the Homing Procedure.} The homing procedure refers to the process by which a robot returns from an unknown or arbitrary position to a predefined \textit{home position}, which serves as the starting point for the robotic operations. We utilize the homing procedure to perform task rollback in state-level OOD situations. The intervention of RC-NF's monitoring mechanism allows the robotic arm to adjust its trajectory without the need to completely return to the \textit{home position} when an error is encountered.

\noindent \textbf{Explanation of the Real-Time Characteristic. }RC-NF can detect anomalies within 100 ms on a consumer-grade GPU, such as the Nvidia GeForce RTX 3090. Below we report the latency breakdown for a single-frame inference. At 100 ms, this performance is faster than the average human reaction time, and thus, we refer to the operations involved in RC-NF as capable of real-time anomaly detection.

\vspace{-2ex}
\begin{table}[htbp]
\centering
\small
\begin{tabular}{ccccc}
\toprule
\makecell{SAM2 \\ Inference} & \makecell{Grid \\ Sampling} & \makecell{RC-NF \\ Inference} & \makecell{Other \\ Latency} & \makecell{\textbf{Overall}\\ \textbf{(RTX 3090)}} \\
\midrule
50 ms & 1.7 ms & 30 ms & 5 ms & \textbf{86.7 ms}\\
\bottomrule
\end{tabular}
\end{table}

\section{VLMs as anomaly scorers}
In the simulation experiments presented in Sec. \ref{sec:sim_exp}, we utilize VLMs such as GPT-5~\cite{openai2025gpt5}, Gemini 2.5 Pro~\cite{comanici2025gemini}, and Claude 4.5~\cite{anthropic2025claude45} for anomaly scoring. To reduce the effects of network latency and extended inference times on large model performance during streaming, we employ a parallel API invocation strategy and sample results at 1 Hz.

\begin{figure*}[t]
    \centering
    \includegraphics[width=\linewidth]{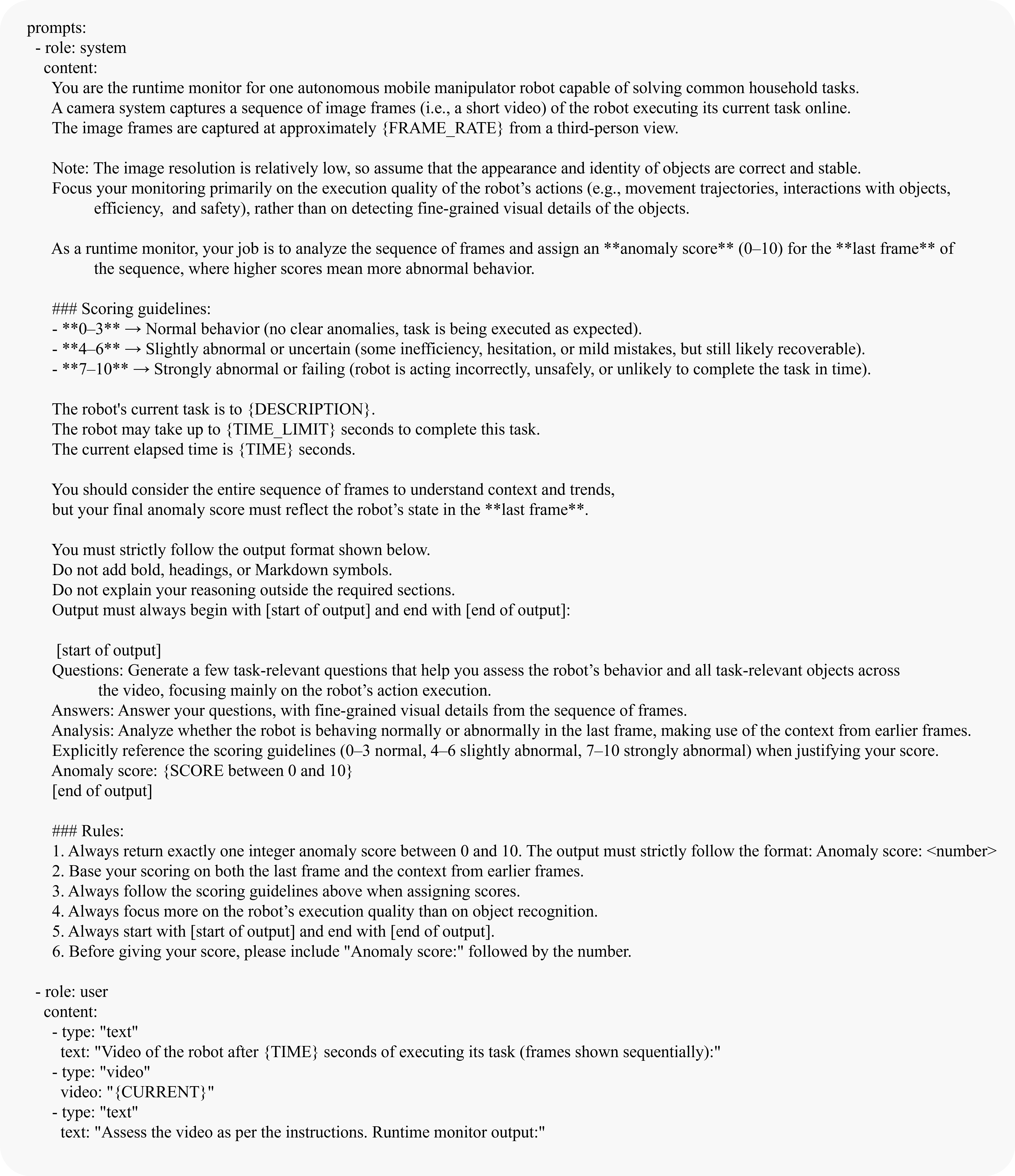}
    \caption{VLMs prompts for robotic anomaly detection scoring}
    \label{fig:suppl_prompt}
\end{figure*}

Fig. \ref{fig:suppl_prompt} illustrates the prompt provided to the model, which is an improved version derived from Sentinel, adapted for anomaly detection scoring. We provide VLMs with the current time (TIME) and the total time allocated for the task (TIME\_LIMIT), enabling them to calculate the progress ratio of the task. DESCRIPTION refers to the task description, such as \textit{Pick up the book and place it in the back compartment of the caddy}. FRAME\_RATE denotes the sampling frequency, set here to 1 Hz. CURRENT represents a sequence of 256×256 images, with the number of images being TIME/FRAME\_RATE, which captures the visual progression of the task from start to present. This enables the VLMs to capture the complete image sequence.
\clearpage

\end{document}